\documentclass[10pt,twocolumn,letterpaper]{article}

\usepackage{iccv}
\usepackage{times}
\usepackage{epsfig}
\usepackage{graphicx}
\usepackage{amsmath}
\usepackage{amssymb}
\usepackage{multirow}
\usepackage{color}

\usepackage{cuted}
\usepackage{capt-of}

\usepackage[pagebackref=true,breaklinks=true,letterpaper=true,colorlinks,bookmarks=false]{hyperref}

\iccvfinalcopy 

\ificcvfinal\fi

\usepackage[T1]{fontenc}
\usepackage[utf8]{inputenc}
\usepackage{authblk}

\author[1,*]{Ruizhi Shao}
\author[2,*]{Gaochang Wu}
\author[1]{Yuemei Zhou}
\author[3]{Ying Fu}
\author[1]{Yebin Liu}
\affil[1]{Tsinghua University}
\affil[2]{Northeastern University}
\affil[3]{Beijing Institute of Technology}

\begin{document}

\title{LocalTrans: A Multiscale Local Transformer Network for Cross-Resolution Homography Estimation}


\maketitle
\ificcvfinal\thispagestyle{empty}\fi

\begin{strip}\centering
\vspace{-16mm}
\includegraphics[width=\textwidth]{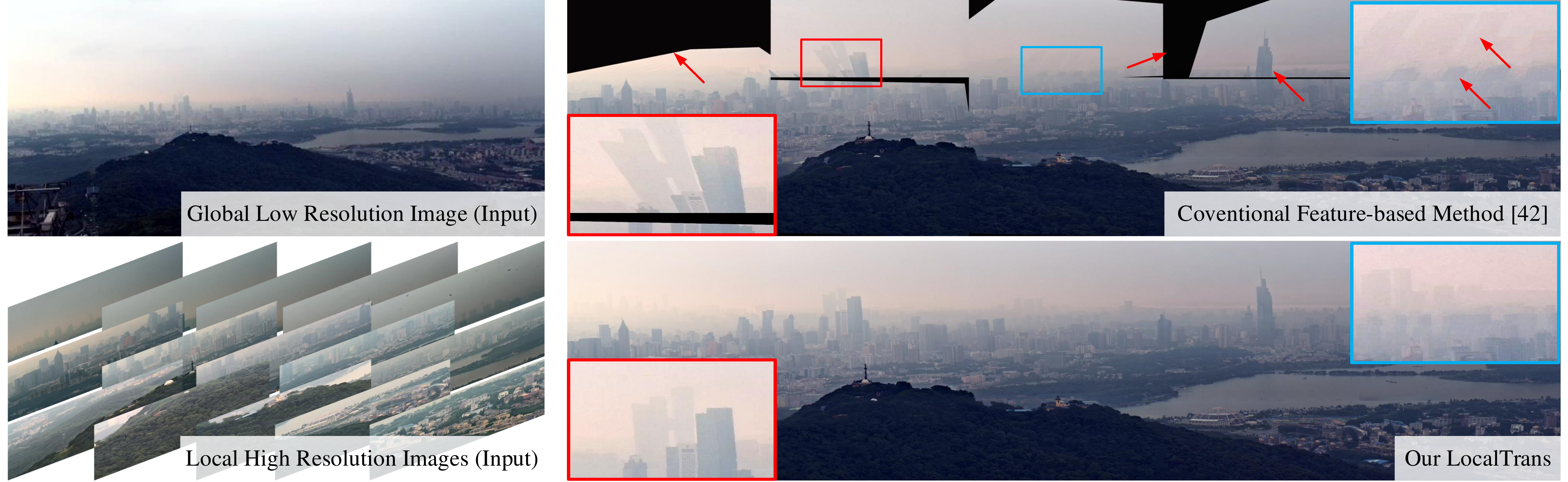}
\vspace{-4mm}
\captionof{figure}{We present a multiscale local transformer network, dubbed \textit{LocalTrans}, for homography estimation. The proposed LocalTrans network achieves accurate homography estimation on challenging real-captured cross-resolution cases under resolution gap up to $10\times$. By warping each of the local high-resolution image on a global low-resolution image using the estimated homography matrix, we achieve artifact-free stitching on this challenging case, substantially outperforms feature-based homography estimation in \cite{yuan2017multiscale}.}
\vspace{-2mm}
\label{teaser}
\end{strip}
\let\thefootnote\relax\footnotetext{* Equal contribution}
\begin{abstract}
\vspace{-4mm}
Cross-resolution image alignment is a key problem in multiscale gigapixel photography, which requires to estimate homography matrix using images with large resolution gap. Existing deep homography methods concatenate the input images or features, neglecting the explicit formulation of correspondences between them, which leads to degraded accuracy in cross-resolution challenges. In this paper, we consider the cross-resolution homography estimation as a multimodal problem, and propose a local transformer network embedded within a multiscale structure to explicitly learn correspondences between the multimodal inputs, namely, input images with different resolutions. The proposed local transformer adopts a local attention map specifically for each position in the feature. By combining the local transformer with the multiscale structure, the network is able to capture long-short range correspondences efficiently and accurately. Experiments on both the MS-COCO dataset and the real-captured cross-resolution dataset show that the proposed network outperforms existing state-of-the-art feature-based and deep-learning-based homography estimation methods, and is able to accurately align images under $10\times$ resolution gap.

\end{abstract}

\section{Introduction}
The rapidly development of multiscale gigapixel photography~\cite{brady2012multiscale,yuan2017multiscale,zhang2020multiscale} brings large-scale, long-term and immersive visual experience. It synthesizes a single ultra-high-resolution image through aligning plenty of high-resolution local-views with a low-resolution global-view. In multiscale gigapixel photography, the large resolution gap between two views, namely cross-resolution, puts forward a new challenge to traditional homography estimation task. Homography estimation is defined as the estimation of the projection mapping between two views on the same plane in 3D space, which usually consists of three steps: feature extraction using SIFT~\cite{lowe2004distinctive} or SURF~\cite{bay2006surf}, correspondence matching, and homography matrix estimation based on the RANSAC~\cite{fischler1981random} or a direct linear transform. It relies on dense features with the same resolution to achieve an accurate estimation, thus, usually fail in solving the cross-resolution problem.

Inspired by the success of deep learning, deep homography methods based on convolutional neural network are studied to deal with challenging scenes. The pioneer deep homography method proposed by DeTone~\etal~\cite{DeTone.2016deephomography} implements the estimation of homography matrix with a typical VGG-net~\cite{simonyan2014very}, which extracts correspondences from the concatenated image pair. Based on this pioneer work, Le~\etal~\cite{Le.2020dynamic} propose a multiscale strategy to progressively estimate the homography via network cascade. However, since the input views are concatenated and downsampled together, simply applying the multiscale strategy cannot solve the cross-resolution problem. A recent approach by Zhang~\etal~\cite{Zhang.2019contentdeephomoraphy} proposed to extract features from input images separately with shared convolution layers. While the network directly concatenates the features in the following layer, which can be equivalent to concatenating the input images at the very beginning.

In this paper, we present a novel multiscale local transformer network, which we dubbed \textit{LocalTrans}, to solve the cross-resolution problem in homography estimation. The transformer structure~\cite{vaswani2017attention} has made a great success in learning the interaction between multimodal inputs~\cite{kant2020spatially,paraskevopoulos2020multimodal,yu2020multimodal} in the field of natural language processing and visual question answering. We therefore take a look at the cross-resolution problem through the lens of ``multimodal'', and employ the transformer structure to explicitly capture correspondences through the correlation of the cross-resolution images in the feature space.


However, the vanilla transformer structure introduced in~\cite{vaswani2017attention} brings high GPU memory and computational costs due to the outer product between high-dimensional matrices. To achieve a fast and accurate homography estimation, we introduce a local transformer and embed it within a multiscale structure. More specifically, we design a local convolution-based operation in the proposed local transformer, which applies a specific kernel to each position of the high-level feature to efficiently capture a local attention. Then the local transformer is deployed in each level of the multiscale structure, enabling the network to capture correspondences with a long-short range attention. The combination of the local transformer and the multiscale structure is significantly faster than the global attention mechanism in the vanilla transformer~\cite{vaswani2017attention}. But most importantly, the proposed LocalTrans network shows a superiority to the vanilla transformer with the same backbone in the homography estimation task.

Benefiting from the combination of the local transformer layer and the multiscale structure, the proposed LocalTrans network outperforms the state-of-the-art homography estimation methods in terms of PSNR and corner error on the MS-COCO dataset~\cite{lin2014microsoft}. Moreover, we demonstrate that the LocalTrans network highlights a superior performance on challenging real-captured cross-resolution cases under resolution gap up to $10\times$, and further apply it to multiscale gigapixel photography, see Fig.~\ref{teaser}. The main contributions are summarized as
\begin{itemize}
    \item We propose to solve the cross-resolution problem in homography estimation using the transformer structure by explicitly capturing the correspondences between the inputs.
    \item We design a novel local transformer layer embedded within multiscale structure, which is able to capture correspondences with a long-short range attention. Experiments demonstrate that the proposed structure outperforms the global attention mechanism.
    \item The proposed local transformer has significantly faster speed and lower GPU memory cost compared with the vanilla transformer structure, achieving real-time homography estimation at 60fps (please see Table~\ref{tab:memoryspeed}).
\end{itemize}

\begin{figure*}[ht]
    \centering
    \includegraphics[width=1\linewidth]{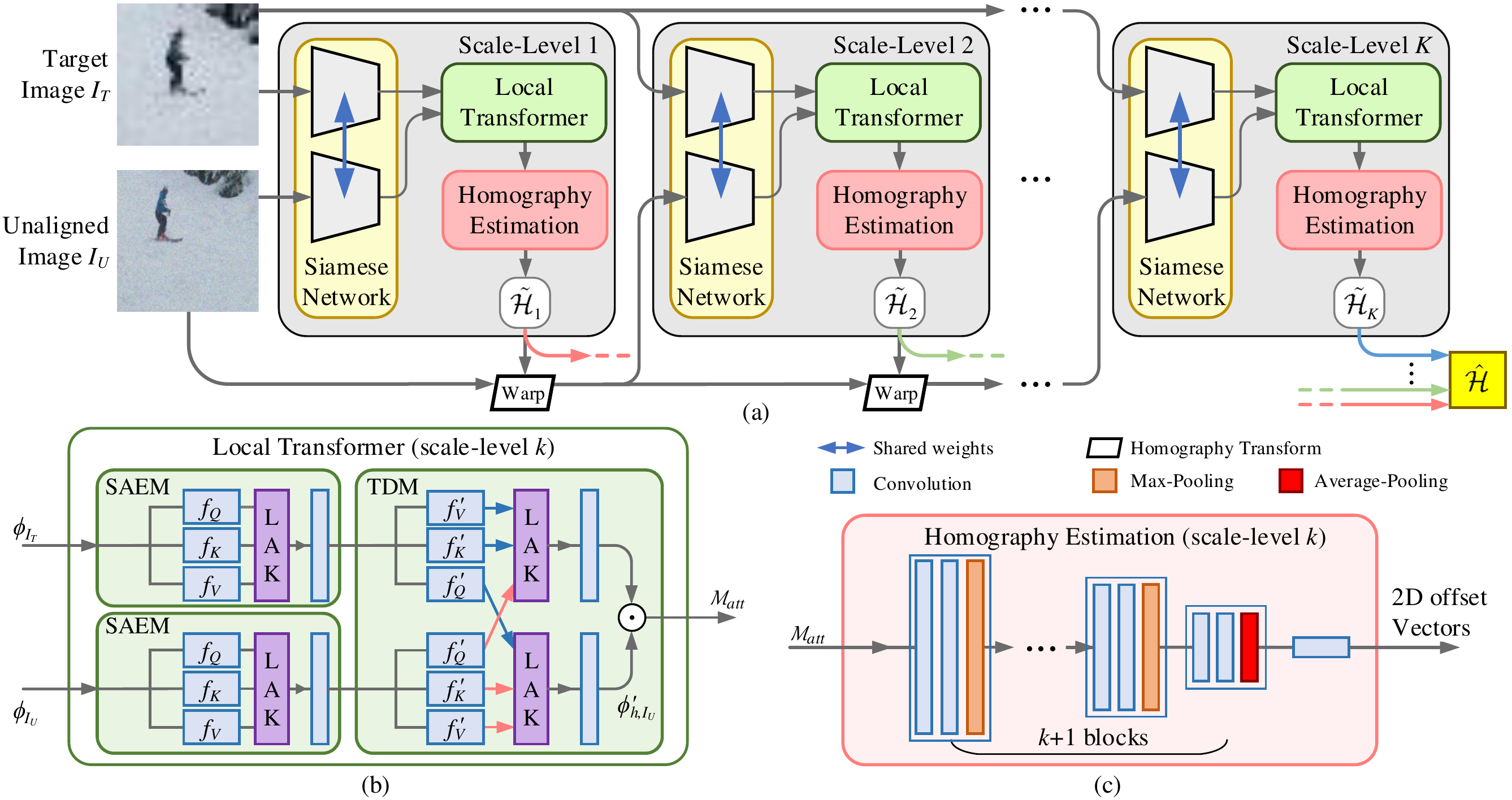}
    \vspace{-4mm}
    \caption{Architecture of the proposed LocalTrans network for homography estimation. (a) Overall structure of the LocalTrans; (b) Architecture of the local transformer that captures correspondences in different scales via a local self-attention encoder module (SAEM) and a local transformer decoder module (TDM); (c) Architecture of the homography estimation module that adopts local attention maps and high-level feature as input to estimate homography matrices from coarse-to-fine.}
    \vspace{-4mm}
    \label{fig:pipeline}
\end{figure*}

\section{Related Work}
In this section, we review topics on homography estimation, cross-resolution image alignment and attention mechanism that are the most relevant to our work.

\textbf{Feature-based Homography Estimation.}
Methods in this category utilize feature points extracted from the image pair to obtain a set of feature correspondences. Then a homography is estimated based on the direct linear transform or the robust fitting algorithms such as RANSAC~\cite{fischler1981random}. The accuracy of homography estimation depends on the quality of the detected image features. Traditional feature detectors, such as SIFT~\cite{lowe2004distinctive}, SURF~\cite{bay2006surf}, ORB~\cite{rublee2011orb}, are able to detect reasonable keypoints, which are robust to lighting, blur and perspective distortion. Recently, several deep learning-based feature extraction methods are also developed, e.g., LFNet~\cite{ono2018lf} and ASLFeat~\cite{luo2020aslfeat}, and reach a higher matching accuracy. Despite existing feature-based homography estimation methods can be robust to illuminance changes and foggy inputs, they
often fail in cross-resolution cases.

\textbf{Deep Homography Estimation.}
Deep learning-based homography estimation is first proposed by~\cite{DeTone.2016deephomography} using VGG-net~\cite{simonyan2014very} as backbone, which is more robust compared to traditional feature-based methods. To improve the generalization capacity on real data, Nguyen~\etal~\cite{nguyen2018unsupervised} proposed an unsupervised learning method by minimizing a pixel-wise intensity error metric instead of the regression loss of homography matrix. To address the potential large motion problem, Le~\etal~\cite{Le.2020dynamic} proposed to use a cascade strategy to estimate the motion mask and the homography matrix in a coarse-to-fine manner. The above methods implicitly estimate the correlation between the two views by concatenating the images along the channel dimension. Alternatively, Zhang~\etal~\cite{Zhang.2019contentdeephomoraphy} proposed to use a feature extractor with shared weights to extract image features separately, and directly concatenate the features in the following homography estimator network. However, this architecture is equivalent to concatenating the input images at the beginning of the network, and thus, fail to address the cross-resolution problem, please refer to the comparison in Fig. \ref{fig:result_synthesis} and Table.\ref {tab:result2}.

\textbf{Cross-Resolution Image Alignment.}
Cross-resolution image alignment is an open problem in cross-scale stereo~\cite{zhou2020cross}, hybrid light field imaging~\cite{2014Improving,zheng2017learningcross}, multiscale gigapixel videography~\cite{yuan2017multiscale}, etc. Commonly a pixel-to-pixel warping field is estimated and applied for registration between images with different resolutions. For example, Zhao~\etal~\cite{zhao2018cross} presented a disparity estimation and refinement method for reconstructing high resolution light field in a hybrid light field imaging system. Zheng~\etal~\cite{Zheng.2018crossnet} further proposed a deep learning-based optical flow estimation and image fusion method in a coarse-to-fine manner. To synthesize a multiscale gigapixel video, Yuan~\etal~\cite{yuan2017multiscale} proposed an iterative feature matching and warping method to perform global and mesh-based homography estimations. In this paper, we investigate the cross-resolution problem in the homography estimation task by using transformer structure to pay attention to the correspondences across different resolution inputs. 

\textbf{Attention Mechanism.}
Attention was built to imitate the mechanism of human perception that mainly focuses on the salient part~\cite{itti1998model,rensink2000dynamic,corbetta2002control}. Vaswani~\etal~\cite{vaswani2017attention} indicated that the global attention mechanism is able to solve the long term dependency problem even without the backbone of a convolutional or a recurrent network. Wang~\etal~\cite{wang2018non-local} introduced a self-attention to capture long-range dependencies (i.e., correspondences) by using matrix multiplication between reshaped feature maps. Alternative to the above global attention, many researches also investigated local attention approaches that focus on short-range dependencies~\cite{gulati2020conformer,luong2015effective,sperber2018self,yang2018modeling,wang2017residual}. For example, Sperber~\etal~\cite{sperber2018self} introduces a soft Gaussian bias and a hard mask that is non-zero in a local region to control the context range attended by the network. Woo~\etal~\cite{woo2018cbam} introduced a spatial attention module by using a convolution layer to extract inter-spatial relationship in a feature map. 

Recently, numerous literatures show that the attention mechanism is efficient to obtain correlation across multimodal inputs, e.g., for visual question answering~\cite{younes2017mutan,kim2018bilinear,yu2017multi}, video description~\cite{hori2017attention} and texture transferring~\cite{yang2020learning}. We therefore apply the attention mechanism to explicitly capture the correspondences across the input images, especially cross-resolution images, for homography estimation.

\section{Methodology}
\subsection{Overview}\label{sec:overview}
In this paper, we introduce a novel deep homography estimation network with a multiscale local transformer structure, dubbed \textit{LocalTrans}. The proposed LocalTrans network first applies a deep siamese network (i.e., an image encoder with shared weights) to extract features $\phi_{I_T}$ and $\phi_{I_U}$ from the target image $I_T$ and the unaligned image $I_U$, respectively. Two convolution layers followed by a $2\times2$ max-pooling layer constitute a basic block in the deep siamese network. Therefore, we construct features with different scales in the multiscale structure by controlling the block number of the deep siamese network in each scale-level. In concrete details, we use $K-k+1$ blocks to construct feature maps $\phi^{(k)}_I$ of shape $H_k\times W_k$ in scale level $k$ ($k=[1,2,\dots,K]$), where $H_k=\frac{H}{2^{K-k+1}}$, $W_k=\frac{W}{2^{K-k+1}}$, and $H$ and $W$ are the height and width of the input images.

Different from existing deep homography methods that simply concatenate the images or features, we then explicitly formulate correspondences between the features $\phi_{I_T}$ and $\phi_{I_U}$ by using the transformer structure (Sec. \ref{sec:transformer}). In each scale-level, a homography estimation module (Sec. \ref{sec:homography}) is adopted to estimate a homography matrix $\tilde{\mathcal{H}}_k$ based on the attention map and feature maps. Then the unaligned image $I_U$ is warped according to the homography matrix $\tilde{\mathcal{H}}_k$ and then fed into the next scale-level (please refer to Fig. \ref{fig:pipeline} (a)).

\begin{figure}
    \centering
    \includegraphics[width=1\linewidth]{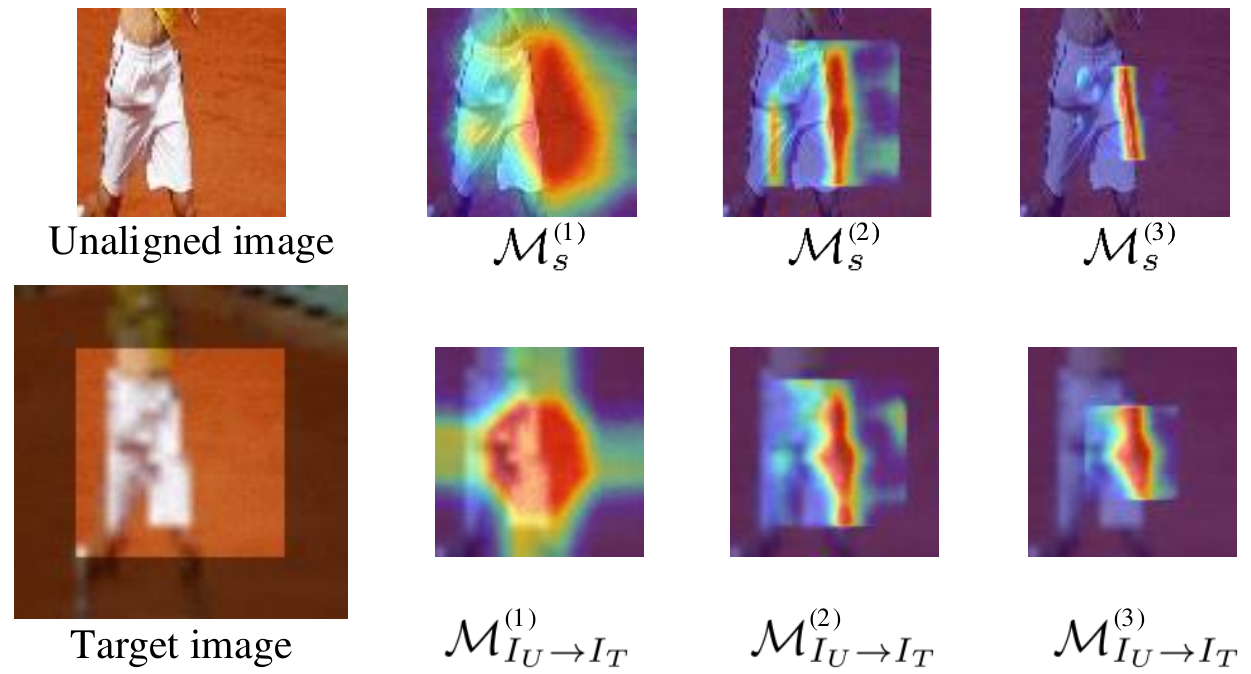}
    \vspace{-4mm}
    \caption{Visualization of the attention map. Top: self-attention map $\mathcal{M}^{(k)}_{s,I_U}$; Bottom: cross-attention map $\mathcal{M}^{(k)}_{I_U\rightarrow I_T}$.}
    \label{fig:att_visual}
    \vspace{-4mm}
\end{figure}

\subsection{Multiscale Local Transformer Network}\label{sec:transformer}
In this section, we introduce the proposed local transformer incorporating a multiscale structure. A straightforward option to achieve a long-short range perception of correspondence between features is to implement the transformer structure~\cite{vaswani2017attention} after the siamese network. However, the vanilla transformer structure brings high GPU memory and computational costs when processing high-dimensional features. To accelerate the transformer, we replace the global attention in the vanilla transformer structure by designing a novel local attention kernel (Sec. \ref{sec:local transformer}) and combining it with a multiscale structure. Despite the designed local transformer kernel only captures correspondences in a limited range in the low scale-level, the multiscale structure enables the network to perceive the correspondence in a long-short range manner.

\vspace{-3mm}
\subsubsection{Transformer Structure}
\vspace{-2mm}
The detailed architecture of the transformer structure in each scale-level is shown in Fig. \ref{fig:pipeline}(b). The inputs of the transformer are two features $\phi^{(k)}_{I_T}$ and $\phi^{(k)}_{I_U}$ output by the deep siamese network. Two modules, Self-Attention Encoder Module (SAEM) and Transformer Decoder Module (TDM), are employed to exploit internal relations within the feature maps via self-attention and to capture the correspondences across the two features from the multimodal inputs via the cross-attention, respectively.

\textbf{Self-Attention Encoder Module (SAEM).} The SAEM first applies three $1\times1$ convolution layers $f_Q(\phi_I)$, $f_K(\phi_I)$ and $f_V(\phi_I)$ without activation function to encode the input image feature $\phi_I$ ($\phi^{(k)}_{I_T}$ or $\phi^{(k)}_{I_U}$) into features $\phi_{Q,I}$, $\phi_{K,I}$, $\phi_{V,I}$ of shape $C\times H_k\times W_k$, where $C$ denotes the channel number. Then the self-attention result in the SAEM is computed as follows
\vspace{-2mm}
\begin{equation}\label{eq:sam}
\begin{split}
\mathcal{M}_s &= \sigma(\frac{\phi_{Q,I}\odot \phi_{K,I}}{\sqrt{C}}),\\
\phi_{h,I} &= \mathcal{M}_s\otimes \phi_{V,I},
\end{split}
\end{equation}
where $\sigma$ denotes the softmax function, and $\odot$ and $\otimes$ denote the operations in the designed local transformer structure, which will be described in Sec. \ref{sec:local transformer}. The tensor $\mathcal{M}_s$ is usually interpreted as a self-attention map.

Since $\phi_{Q,I}, \phi_{K,I}$ and $\phi_{V,I}$ are derived from the same input $\phi_I$, the self-attention mechanism encourages to enhance the edges and corners in the input feature $\phi(I)$. As shown in an example in Fig. \ref{fig:att_visual} (top), the network pays more attention to the edge with the same feature of the center pixel, which is more prominent in attention maps of higher scale-levels (e.g., $\mathcal{M}^{(2)}_s$ and $\mathcal{M}^{(3)}_s$). The final high-level feature $\phi_{s,I}$ output by the SAEM is generated by encoding the self-attention result $\phi_{h,I}$ with a $1\times1$ convolution layer.

\textbf{Transformer Decoder Module (TDM).} In this module, two iterations of cross-attention are used. In the first iteration, we first adopt three $1\times1$ convolution layers $f'_Q(\cdot), f'_K(\cdot), f'_V(\cdot)$ without activation function to encode the high-level features $\phi_{s,I}$ into features $\phi'_{Q,I}$, $\phi'_{K,I}$ and $\phi'_{V,I}$ as those in the SAEM. But different with the self-attention in the SAEM (Eqn. \ref{eq:sam}), we apply a cross-attention mechanism between features from target image $I_T$ and unaligned image $I_U$, denoted as
\vspace{-2mm}
\begin{equation}\label{eq:tdm}
\begin{split}
\mathcal{M}_{I_U\rightarrow I_T} =&\sigma(\frac{\phi'_{Q,I_U}\odot \phi'_{K,I_T}}{\sqrt{C}}), \\
\phi'_{h,I_T} =&\mathcal{M}_{I_U\rightarrow I_T}\otimes \phi'_{V,I_T}, \\
\mathcal{M}_{I_T\rightarrow I_U} =&\sigma(\frac{\phi'_{Q,I_T}\odot \phi'_{K,I_U}}{\sqrt{C}}), \\
\phi'_{h,I_U} =&\mathcal{M}_{I_T\rightarrow I_U}\otimes \phi'_{V,I_U},
\end{split}
\end{equation}
where $\mathcal{M}_{I_U\rightarrow I_T}$ and $\mathcal{M}_{I_T\rightarrow I_U}$ are usually interpreted as cross-attention map, and $\phi'_{h,I_T}$ and $\phi'_{h,I_U}$ are attention-aware features.

In the second iteration, features $\phi'_{h,I_T}$ and $\phi'_{h,I_U}$ are first encoded into $\phi'_{s,I_T}$ and $\phi'_{s,I_U}$ using two $1\times1$ convolution layers, as shown in Fig. \ref{fig:pipeline}(b). Then we compute the attention map $\mathcal{M}_{att}$ using $\phi'_{s,I_T}$ and $\phi'_{s,I_U}$ as inputs
\vspace{-1mm}
\begin{equation}\label{eq:soft_corr}
\mathcal{M}_{att} = \phi'_{s,I_T}\odot \phi'_{s,I_U}.
\vspace{-1mm}
\end{equation}
The attention map $\mathcal{M}_{att}$ is served to estimate the homography matrix, which will be introduced in Sec. \ref{sec:homography}.

The attention mechanism described in Eqn. \ref{eq:tdm} enables an interaction between target image $I_T$ and unaligned image $I_U$ in the feature space, and therefore, captures correspondence information more explicitly than simply feeding the image pair into the network through channel-wise concatenation~\cite{simonyan2014very,DeTone.2016deephomography,nguyen2018unsupervised,Le.2020dynamic}. Besides, the network with the transformer structure is more robust when the input two images have considerably large resolution differences. Fig. \ref{fig:att_visual} (bottom) visualizes the cross-attention map $\mathcal{M}_{I_U\rightarrow I_T}$ for input images under $4\times$ resolution gap. In this cross-resolution case, more attention values (weights) are assigned to the same edge as that in the self-attention map $\mathcal{M}_{s,I_U}$. 

\subsubsection{Local Attention Kernel}\label{sec:local transformer}
\begin{figure}
\centering
    \includegraphics[width=1\linewidth]{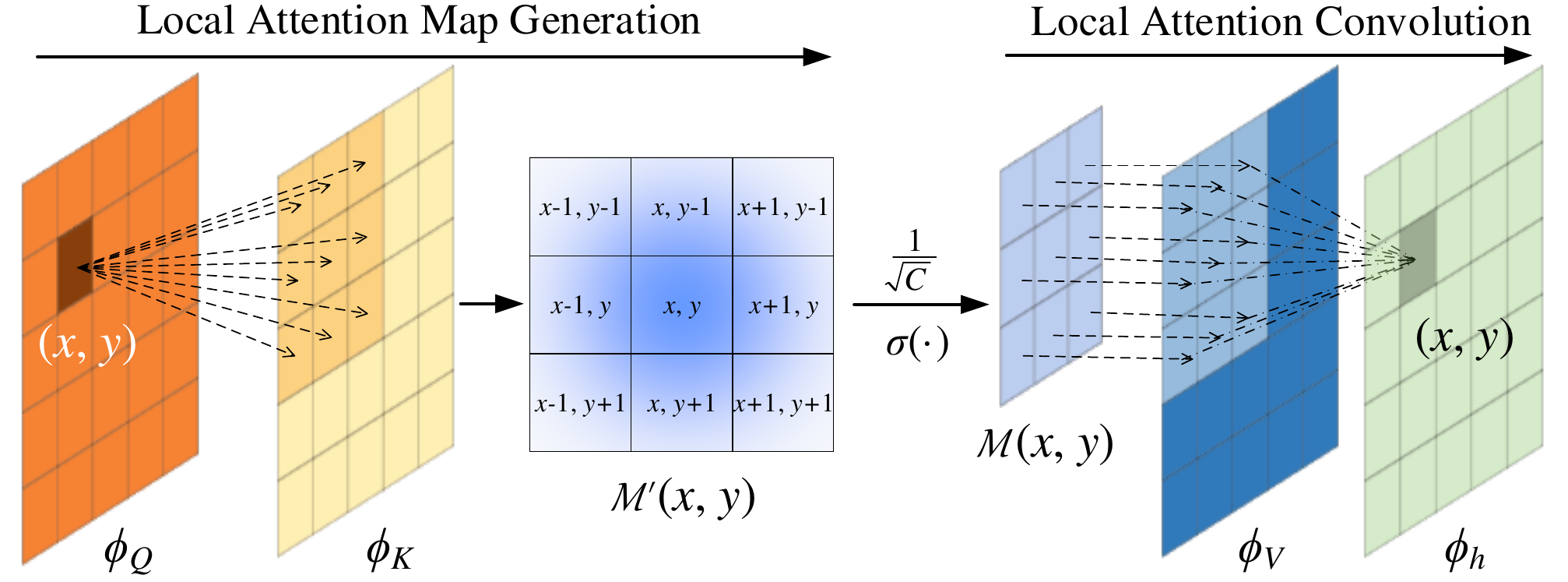}
    \vspace{-4mm}
    \caption{The process of Local Attention Kernel (LAK). In the first step, a 2D slice of the local attention map is generated for each element of feature $\phi_Q$ in position $\textbf{x}$. In the next step, we regard the 2D slice of the local attention map as a convolution kernel for a patch centered at position $\textbf{x}$ in feature $\phi_V$.}
    \vspace{-2mm}
    \label{fig:attention_kernel}
\end{figure}

To accelerate the transformer structure in Sec.~\ref{sec:transformer}, we propose a Local Attention Kernel (LAK) that captures correspondences in a local range, which is inspired by conventional 2D deconvolution (also known as transposed convolution) and convolution. The main difference is that the proposed local transformer applies variable slice in the attention map as the convolution kernel while the traditional 2D convolution or deconvolution adopts a fixed kernel in the feedforward path. In the following, we will introduce the proposed LAK by decomposing the attention mechanism in Eqn. \ref{eq:sam} and \ref{eq:tdm} into two steps, as shown in Fig. \ref{fig:attention_kernel}.

\textbf{Local attention map generation.} Local attention map describes the correspondences between features $\phi_Q$ and $\phi_K$ in a local range, i.e., a squared window. Consider $\phi_Q\in\mathbb{R}^{C\times H_k\times W_k}$, $\phi_K\in\mathbb{R}^{C\times H_k\times W_k}$ and the radius of LAK is $r$. Then for a certain element at position $\textbf{x}=(x, y)$ in $\phi_Q$ will first query the relationship with elements in a local range of $\mathcal{N}(\textbf{x})$ with radius $r$ in $\phi_K$. Suppose the position of an element in $\phi_K$ is $\textbf{u}\in\mathcal{N}(\textbf{x})$, then a local correspondence map $\mathcal{M}'$ could be described as
$$\mathcal{M}'(\textbf{x}, \textbf{u}) = \phi_Q^T(\textbf{x})\phi_K(\textbf{u}),$$
where $\phi_Q^T(\textbf{x})\in\mathbb{R}^{1\times C}$ and $\phi_K(\textbf{u})\in\mathbb{R}^{C\times1}$. The above formulation also explains the operation $\phi_Q\odot\phi_K$ in Eqn. \ref{eq:sam}, Eqn. \ref{eq:tdm} and Eq. \ref{eq:soft_corr}. And the final local attention map $\mathcal{M}$ is
$$\mathcal{M} = \sigma(\frac{\mathcal{M}'}{\sqrt{C}}).$$

The above equation shows that the local attention map $\mathcal{M}$ is a 4D tensor with shape $H_k\times W_k\times(2r+1)\times(2r+1)$, which records the correspondence of each position in $\phi_Q$ with that in $\phi_K$ in a local range. For instance, an element $\mathcal{M}(x,y,u,v)$ records the correspondence between point $\phi_Q(x,y)$ and $\phi_K(x+u,y+v)$, $u,v\in[-r,r]$.

\textbf{Local attention convolution.} This operation uses the 4D local attention map $\mathcal{M}$ and the feature $\phi_V$ to obtain high-level feature $\phi_h$. Consider a 2D slice of $\mathcal{M}$ at a certain position $\textbf{x}=(x, y)$, denoted as $\omega_\textbf{x}$. The feature $\phi_h$ is then obtained by performing the convolution between the feature $\phi_V$ and the 2D slice $\omega_\textbf{x}$
\vspace{-2mm}
$$\phi_h(\textbf{x})=\omega_\textbf{x}*\phi_V(\textbf{x})= \sum_{\textbf{u}\in\mathcal{N}(\textbf{x})} \mathcal{M}(\textbf{x}, \textbf{u}) \phi_V(\textbf{x}+\textbf{u}),\vspace{-2mm}$$
where $\mathcal{M}(\textbf{x}, \textbf{u})\in\mathbb{R}^{1\times 1}$ and $\phi_V(\textbf{x}+\textbf{u})\in\mathbb{R}^{C\times1}$. The above formulation also explains the operation $(\cdot)\otimes\phi_V$ in Eqn. \ref{eq:sam} and Eqn. \ref{eq:tdm}.

\textbf{Discussions.} 
It should be noted that there is a clear difference between the proposed LAK and existing approach with local attention~\cite{gulati2020conformer,sperber2018self,yang2018modeling,wang2017residual,woo2018cbam,wang2020eca}. Existing approaches typically implement local attention by setting an attention bias (a Gaussian bias~\cite{yang2018modeling} or a hard mask that is non-zero in a local region~\cite{sperber2018self}) or convolution layers to perform channel squeezing~\cite{wang2017residual,woo2018cbam,wang2020eca}. In our proposed LocalTrans, the convolution kernel in the LAK, i.e., the 2D slice $\omega_\textbf{x}$, varies with the position $\textbf{x}$, while the kernel in a conventional convolution is fixed in every position. We therefore term the operation as local attention convolution. The most related local attention structure was proposed in~\cite{luong2015effective}, which computes a local weights within a small window and produce feature vector though weighted average. We generalize this concept to 2D space in combination with 2D convolution, making it more suitable for visual tasks.

Comparing with the global attention in the vanilla transformer structure, the proposed LAK effectively reduces the computational complexity from $\mathcal{O}(H_k^2\cdot W_k^2\cdot C)$ to $\mathcal{O}(H_k\cdot W_k\cdot (2r+1)^2\cdot C)$, and the memory usage of the attention map from $H_k^2\cdot  W_k^2$ to $H_k\cdot  W_k\cdot (2r+1)^2$. For a certain scale-level $k\in K$, we set the radius of the LAK $r=k+1$ to encourage the local transformer to notice a longer range of correspondence in the higher levels.

To demonstrate the effectiveness of the proposed local attention, we compare the local transformer against the vanilla transformer with the same deep siamese network and the homography estimation module, as shown in Fig. \ref{fig:eval_vanilla}. The result shows that the proposed LAK with even 1 scale is superior to the global attention in the vanilla transformer structure. In conclusion, the proposed local transformer kernel not only shows high computational efficiency but also has superior performance to the vanilla transformer structure with global perception of correspondences.

\subsection{Homography Estimation Module}\label{sec:homography}
In each scale-level, the homography estimation module applies the attention map $\mathcal{M}_{att}$ in Eqn. \ref{eq:soft_corr} as input. A 8-dimensional vector is obtained by several convolution-pooling blocks to estimate the final homography matrix, as shown in Fig. \ref{fig:pipeline}(c).

More specifically, in a certain scale-level $k$, since the the attention map $\mathcal{M}_{att}$ is a 4D tensor, we first reshape it from $[\frac{H}{2^{K-k+1}}, \frac{W}{2^{K-k+1}},2k+3,2k+3]$ to $[(2k+3)^2, \frac{H}{2^{K-k+1}}, \frac{W}{2^{K-k+1}}]$ (similar to the processing of cost volumn in~\cite{sun2018pwc}) before feeding to the homography estimation module. Then we feed the feature into $k+1$ convolution blocks, where each block contains two $3\times3$ convolution layers and one max-pooling layer. In the last block, the max-pooling layer is replaced by an average-pooling layer for generating a feature of shape $C\times1\times1$. After a 1D convolution layer (kernel size 1), the final output becomes a 8-dimensional vector, denoting the 2D offsets of 4 corner points $\hat{\textbf{c}}^{(k)}_1,\cdots,\hat{\textbf{c}}^{(k)}_4$ in scale-level $k$. We can simply obtain the homography matrix $\tilde{\mathcal{H}}_k$ from the offset vectors and warp the the unaligned image $I_U^{(k)}$ using homography transform to obtain the input for the next level, i.e.,
\vspace{-2mm}$$I_U^{(k+1)} = Warp(I_U^{(k)}, \tilde{\mathcal{H}}_k).\vspace{-2mm}$$
Note that $\tilde{\mathcal{H}}$ in each scale-level represents the homography matrix of the full resolution. Thus, the final homography matrix $\hat{\mathcal{H}}$ is computed directly by accumulating the estimated homography matrix $\tilde{\mathcal{H}}$ in each level as follows
$$\hat{\mathcal{H}} = \tilde{\mathcal{H}}_1 \times \tilde{\mathcal{H}}_2\times \cdots \times \tilde{\mathcal{H}}_K.$$

\subsection{Implementation Details}
In our experiment, we set the number of scale-level $K=3$. Except the $1\times1$ convolution layers $f_Q$, $f_K$, $f_V$, $f'_Q$, $f'_K$ and $f'_V$ in the local transformer and the 1D convolution layer at the end of the homography estimation module, every other 2D convolution layer has a $3\times3$ kernel followed by a batch normalization layer~\cite{ioffe2015batch} and a ReLU activation. More details of the network specification are listed in the supplementary file. For the local transformer layer, we implement the LAK including local attention map generation and local attention convolution in CUDA. To make them differentiable, we also deduce a backward propagation and package it to PyTorch autograd function~\cite{2019PyTorch}.

For the training objective, we use the $L1$ norm of the corner error as the loss function
$\mathcal{L}=\frac{1}{4}\sum_{i=1}^4\|\textbf{c} - \hat{\textbf{c}}_i\|_1$, where $\textbf{c}_i$ and $\hat{\textbf{c}}_i$ are corner point $i$ transformed by the ground-truth homography and the estimated homography, respectively. We only use the MS-COCO dataset~\cite{lin2014microsoft} for the network training. And we follow the same data processing schemes in~\cite{DeTone.2016deephomography,chang2017clkn} to generate image pairs. Moreover, we also add Gaussian noise and randomly adjust brightness, saturation and contrast to increase the robustness of the network.

\section{Experiments}
We compare the proposed LocalTrans network with both feature-based and deep learning-based homography estimation methods. We evaluate the proposed LocalTrans network in two different settings, common data (Sec. \ref{sec:common}) in the MS-COCO dataset~\cite{lin2014microsoft} as that in most deep homography estimation methods, and cross-resolution setting where the target image has lower resolution.

We have two kinds of datasets for the cross-resolution setting. The first is synthesized cross-resolution data (Sec. \ref{sec:cross1}), in which the target images are downsampled using bicubic interpolation with factors $4\times$ and $8\times$. The second is optical zoom-in cross-resolution data (Sec. \ref{sec:cross2}), which we apply multiscale gigapixel dataset from Yuan~\etal~\cite{yuan2017multiscale} and cross-resolution stereo dataset from Zhou~\etal~\cite{zhou2020cross}. In the cross-resolution setting, the proposed LocalTrans network as well as the baseline networks are re-trained.

\subsection{Data in Common Setting}\label{sec:common}
\begin{figure}
    \centering
    \includegraphics[width=\linewidth]{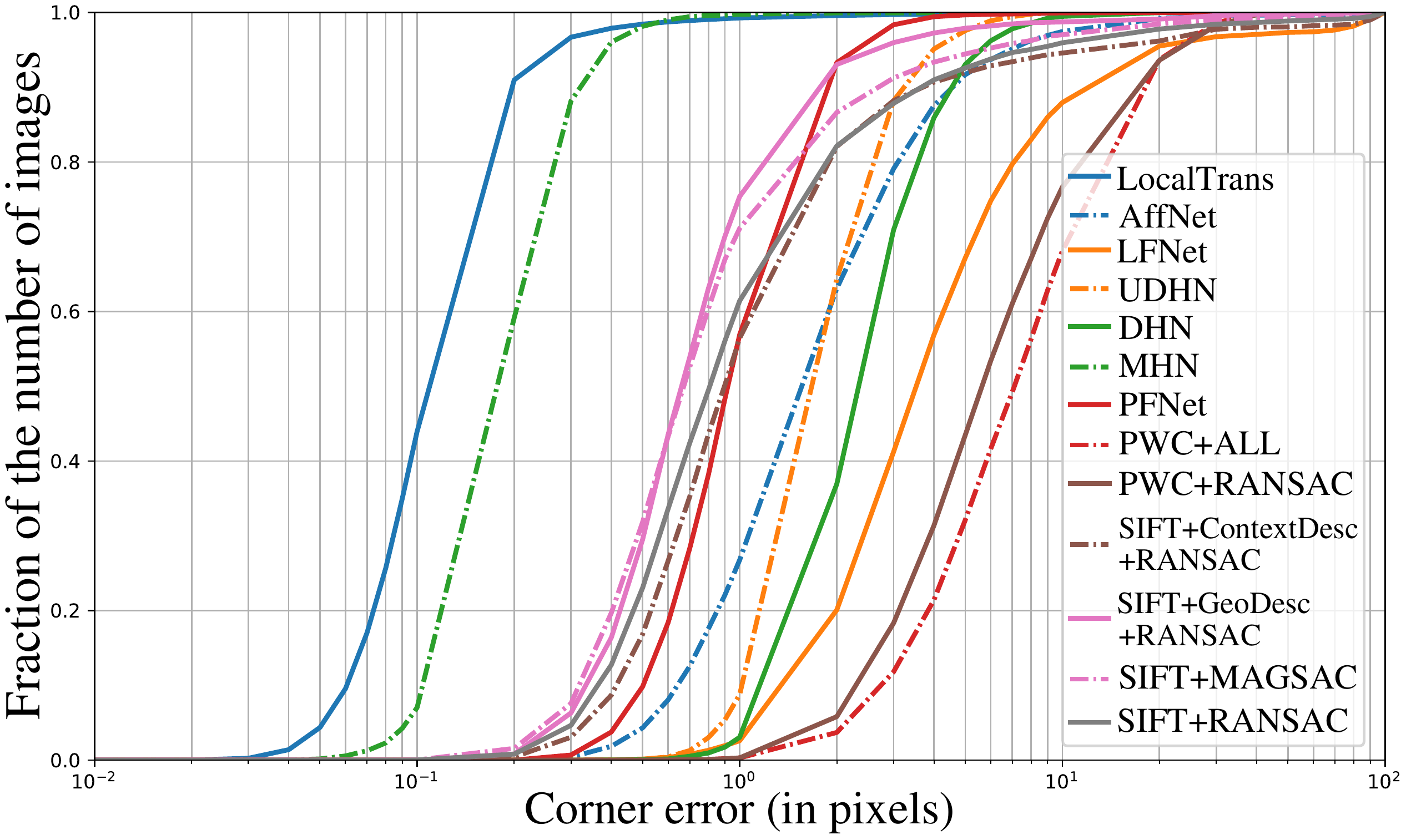}
    \vspace{-5mm}
    \caption{Evaluations in common setting on the MS-COCO.}
    \label{fig:eval_coco}
\end{figure}

\begin{figure}
    \centering
    \includegraphics[width=\linewidth]{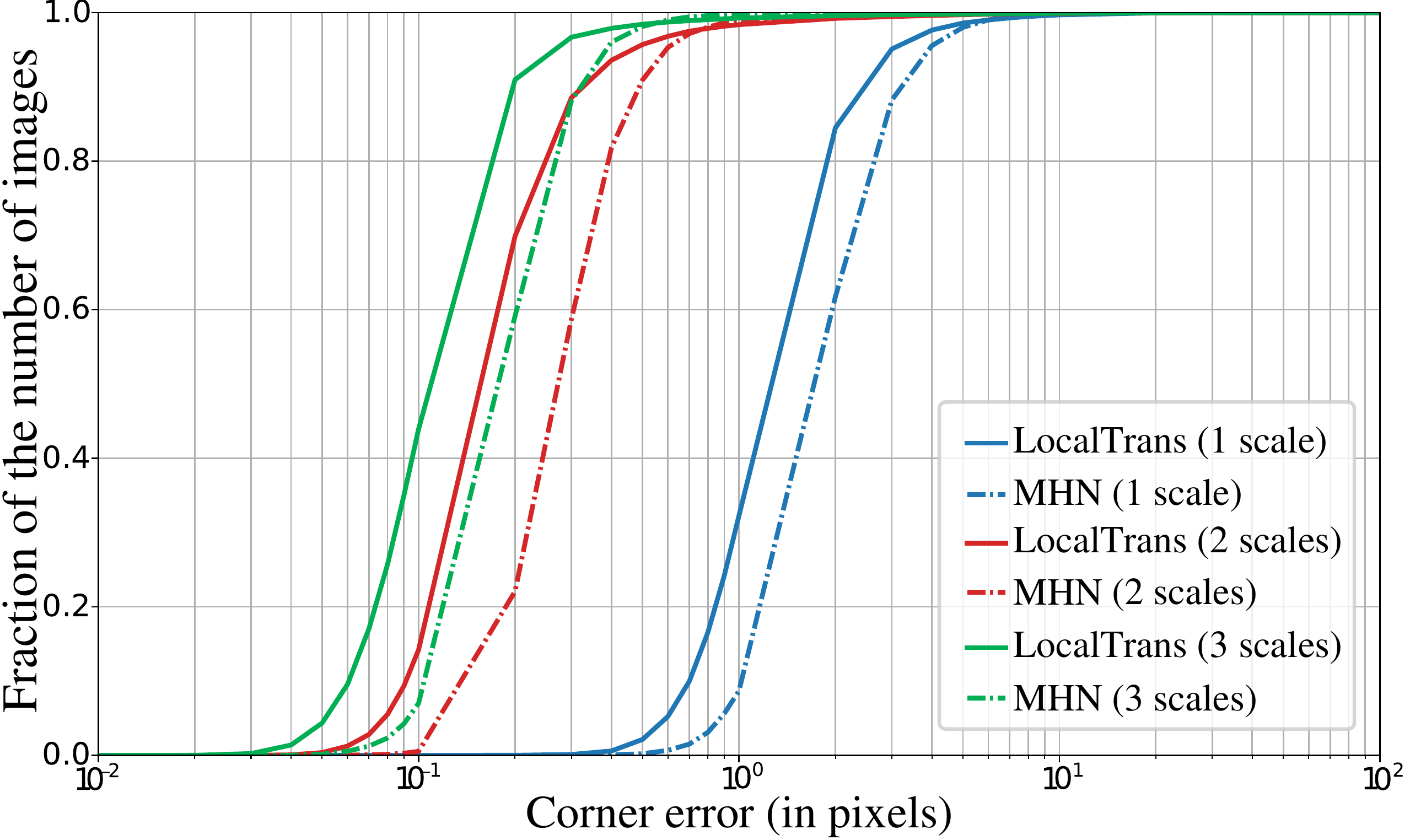}
    \vspace{-5mm}
    \caption{Comparison with MHN \cite{Le.2020dynamic} on the MS-COCO dataset with different numbers of scales.}
    \label{fig:eval_ms}
\end{figure}

We compare our model on the MS-COCO dataset~\cite{lin2014microsoft} in common setting with the following baseline methods, AffNet~\cite{mishkin2018aff}, LFNet~\cite{ono2018lf}, DHN\cite{DeTone.2016deephomography}, UDHN by Zhang~\etal~\cite{Zhang.2019contentdeephomoraphy}, MHN~\cite{Le.2020dynamic}, PFNet~\cite{zeng2018rethinking}, PWC~\cite{sun2018pwc}, SIFT +ContextDesc+RANSAC~\cite{luo2019contextdesc}, SIFT+GeoDesc+RANSAC~\cite{luo2018geodesc}, SIFT+MAGSAC~\cite{barath2019magsac}, SIFT+RANSAC~\cite{lowe2004distinctive}.
Fig. \ref{fig:eval_coco} shows that the proposed LocalTrans outperforms feature-based homography estimation methods~\cite{lowe2004distinctive,sun2018pwc,zeng2018rethinking,ono2018lf,luo2019contextdesc,luo2018geodesc,barath2019magsac} and state-of-the-art deep learning-based methods~\cite{DeTone.2016deephomography,mishkin2018aff,Zhang.2019contentdeephomoraphy,Le.2020dynamic} in common setting. We also compare the proposed LocalTran network with a similar multiscale structure-based deep homography network MHN~\cite{Le.2020dynamic} with different numbers of scales on the MS-COCO dataset, as shown in Fig. \ref{fig:eval_ms}. The result shows that the proposed network with the local transformer outperforms the MHN with different scale-levels. Moreover, our network on with 2 scales performs even better than the MHN with 3 scales. The experiment empirically validates that the proposed local transformer is able to capture correspondences more accurately than simply stacking the images~\cite{Le.2020dynamic} or feature maps~\cite{Zhang.2019contentdeephomoraphy} as input.

\begin{table}\footnotesize
    \begin{center}
    \begin{tabular*}{\linewidth}{@{\extracolsep{\fill}}ccccc}
    \hline
         & \multicolumn{2}{c}{Global Transformer} & \multicolumn{2}{c}{Local Transformer} \\
         \cline{2-3}\cline{4-5}
         & Memory & Speed & Memory & Speed \\
    \hline
    $128\times128$ (1 scale) & 52.0M & 152fps & 48.9M & 249fps   \\
    $128\times128$ (2 scales) & 75.0M & 104fps & 50.7M & 203fps \\
    $128\times128$ (3 scales) & 476.9M & 16.4fps & 67.7M & 132fps \\
    \hline
    $256\times256$ (1 scale) & 208.4M & 129fps & 196.9M & 213fps \\
    $256\times256$ (2 scales) & 313.9M & 52.9fps & 204.8M & 173fps\\
    $256\times256$ (3 scales) & 2434M & 4.31fps & 287.0M & 87.7fps\\
    \hline
    \end{tabular*}
    \end{center}
    \vspace{-2mm}
    \caption{Per-image memory consuming and speed comparison between global and local transformer using different sizes of inputs.}
    \vspace{-2mm}
    \label{tab:memoryspeed}
\end{table}

\begin{figure}
    \centering
    \includegraphics[width=\linewidth]{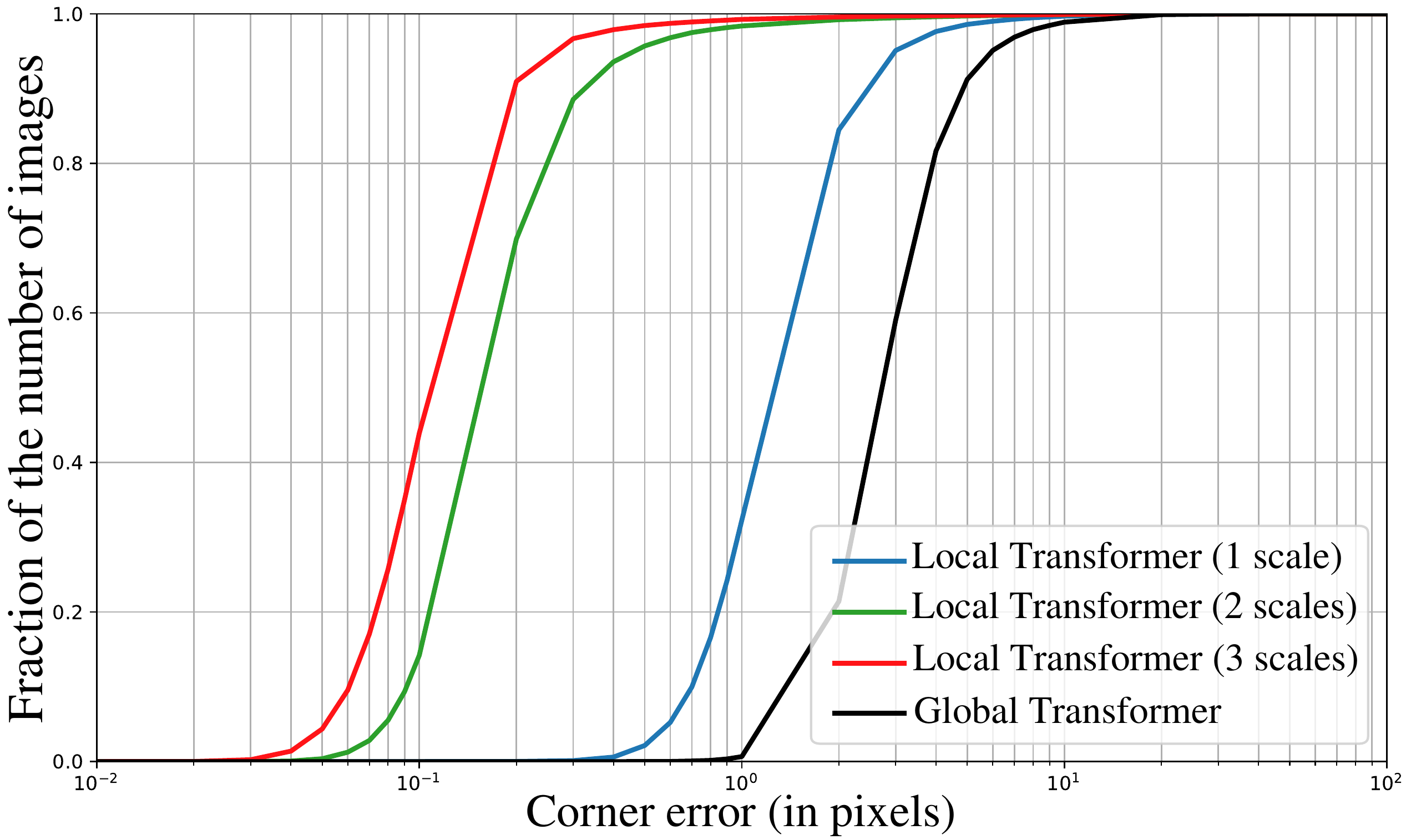}
    \vspace{-5mm}
    \caption{Comparison between the proposed LAK in different scale-levels and the vanilla (global) transformer structure on the MS-COCO dataset~\cite{lin2014microsoft}.}
    \vspace{-1mm}
    \label{fig:eval_vanilla}
\end{figure}

\textbf{Ablation study.} To verify the efficiency of local transformer structure, we replace the proposed LAK with the vanilla (global) transformer in~\cite{vaswani2017attention} while keeping the rest architecture of the network unchanged. This experiment is performed on an Intel(R) Xeon CPU E5-2699 V4 with 16GB memory and a NVIDIA RTX 2080 GPU.
The comparison in Table~\ref{tab:memoryspeed} shows that the proposed LAK has higher computational efficiency in terms of both running time and GPU memory cost than the vanilla transformer. Moreover, results in Fig.~\ref{fig:eval_vanilla} also validate the superior performance compared with a single scale global transformer network. Please refer to the supplementary for more ablation studies.

\begin{figure}
    \centering
    \includegraphics[width=\linewidth]{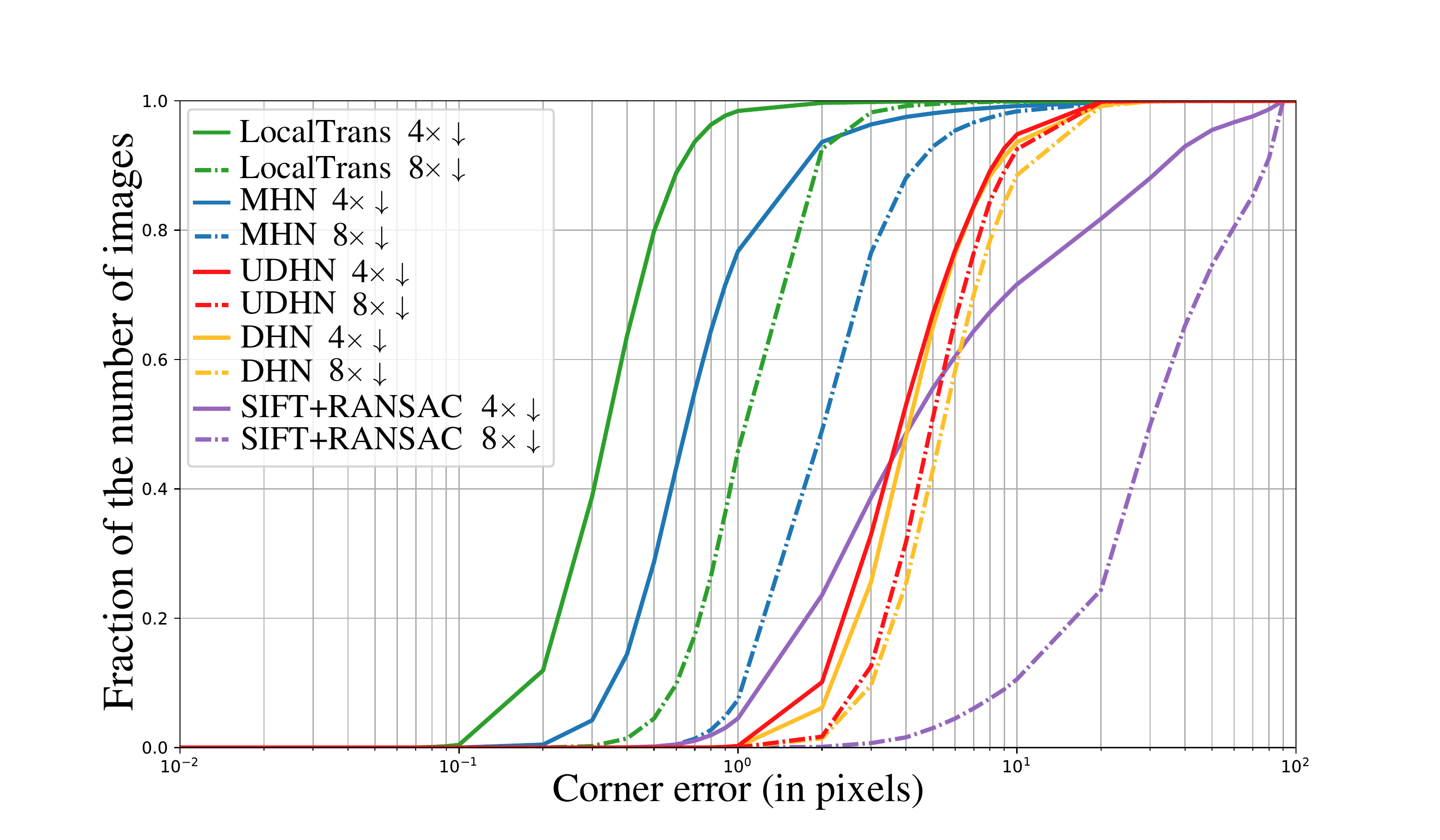}
    \vspace{-5mm}
    \caption{Evaluation on the MS-COCO dataset under $4\times$ and $8\times$ resolution gaps.}
    \vspace{-2mm}
    \label{fig:result_synthesis}
\end{figure}

\subsection{Synthesized Cross-Resolution Data}\label{sec:cross1}
We compare our model on synthesized cross-resolution dataset with 6 baseline methods, a coventional feature-based method, SIFT+RANSAC~\cite{lowe2004distinctive}, three deep learning-based methods, DHN~\cite{DeTone.2016deephomography}, UDHN~\cite{Zhang.2019contentdeephomoraphy} and MHN~\cite{Le.2020dynamic}, as well as two state-of-the-art Reference-based Super-Resolution (RefSR) methods, SRNTT~\cite{zhang2019image} and TTSR~\cite{yang2020learning}.


\begin{table}\footnotesize
    \begin{center}
    \begin{tabular}{lcccc}
    \hline
    \multirow{2}*{Method}     & \multicolumn{2}{c}{$4\times\downarrow$} & \multicolumn{2}{c}{$8\times\downarrow$} \\
    \cline{2-3}\cline{4-5}
         & PSNR & SSIM & PSNR & SSIM \\
    \hline
    SIFT+RANSAC~\cite{lowe2004distinctive} & 19.32 & 0.802 & 11.93 & 0.64 \\
    DHN~\cite{DeTone.2016deephomography} & 19.64 & 0.818 & 17.04 & 0.762 \\
    UDHN~\cite{Zhang.2019contentdeephomoraphy} & 21.44 & 0.864 & 18.78 & 0.834 \\
    MHN~\cite{Le.2020dynamic} & 25.42  & \textcolor[rgb]{0,0,1}{0.951} & \textcolor[rgb]{0,0,1}{20.24} & \textcolor[rgb]{0,0,1}{0.860} \\
    SRNTT~\cite{zhang2019image}  & 27.06  & 0.901 & - & - \\
    TTSR~\cite{yang2020learning} & \textcolor[rgb]{0,0,1}{27.89} & 0.915 & - & - \\
    LocalTrans & \textcolor[rgb]{1.00,0.00,0.00}{30.17} & \textcolor[rgb]{1.00,0.00,0.00}{0.981} & \textcolor[rgb]{1.00,0.00,0.00}{24.12} & \textcolor[rgb]{1.00,0.00,0.00}{0.930} \\
    \hline
    \end{tabular}
    \end{center}
    \vspace{-2mm}
    \caption{Numerical comparison (PSNR/SSIM) in different cross-scale setings on the MS-COCO datasets.}
    \vspace{-3mm}
    \label{tab:result2}
\end{table}

\begin{figure*}
    \centering
    \includegraphics[width=\linewidth]{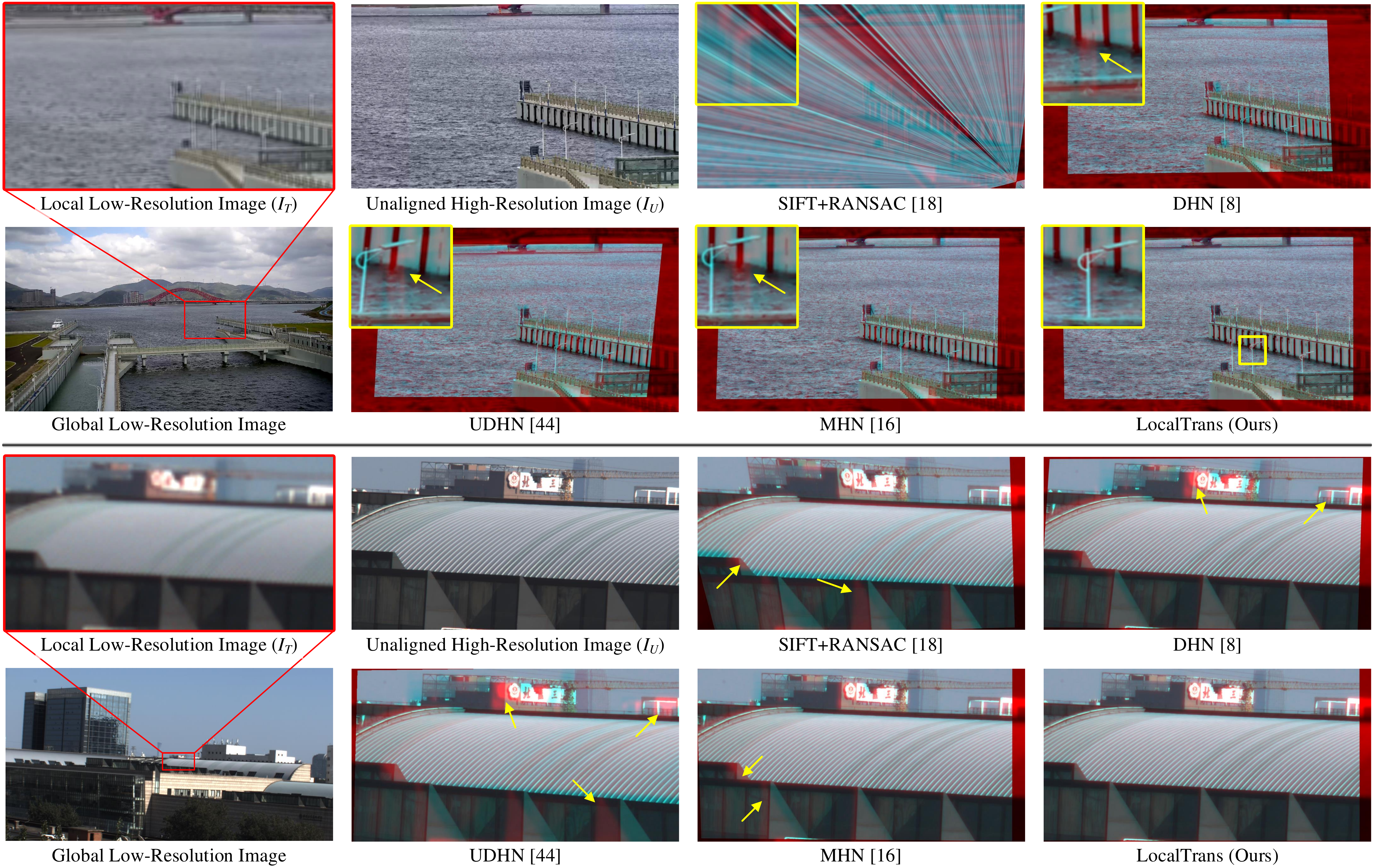}
    \vspace{-4mm}
    \caption{Visual evaluation on the multiscale gigapixel dataset~\cite{yuan2017multiscale} (top, $6\times$) and the cross-resolution stereo dataset~\cite{zhou2020cross} (bottom, $10\times$). We mix the GB channels of the aligned image and the R channel of the target image. The misaligned pixels appear as red or green ghosts.}
    \vspace{-2mm}
    \label{fig:compare_giga}
\end{figure*}

Fig.~\ref{fig:result_synthesis} shows the quantitative comparison on the MS-COCO dataset under $4\times$ and $8\times$ resolution gaps. The proposed LocalTrans network demonstrates superior performance on cross-resolution cases comparing with the conventional feature-based method, SIFT+RANSAC~\cite{lowe2004distinctive}, and two deep learning-based methods, DHN~\cite{DeTone.2016deephomography} and MHN~\cite{Le.2020dynamic}. The numerical comparison in Table~\ref{tab:result2} shows that the proposed LocalTrans network significantly outperforms the deep learning-based methods, DHN~\cite{DeTone.2016deephomography} and MHN~\cite{Le.2020dynamic}, and the RefSR methods, SRNTT~\cite{zhang2019image} and TTSR~\cite{yang2020learning}, which validates the superiority of the proposed local transformer structure in solving the cross-resolution problem.

\subsection{Optical Zoom-in Cross-Resolution Data}\label{sec:cross2}
In this experiment, 4 baseline methods, SIFT+RANSAC \cite{lowe2004distinctive}, DHN~\cite{DeTone.2016deephomography}, UDHN by Zhang~\etal~\cite{Zhang.2019contentdeephomoraphy} and MHN~\cite{Le.2020dynamic} are compared. On the multiscale gigapixel dataset~\cite{yuan2017multiscale}, we then apply the local patches in a $5\times5$ grid of the global low-resolution image to estimate homography matrices separately. To ensure the spatial smoothness between neighbouring patches, we calculate the four corner points of each patch and take an average among its neighbors. The final result is obtained by warping each local high-resolution image to the corresponding grid as in~\cite{yuan2017multiscale}.

Since there is no groundtruth for qualitative evaluation, we only demonstrate the visual comparison on the datasets~\cite{yuan2017multiscale,zhou2020cross}, as shown in Fig.~\ref{fig:compare_giga}. The resolution gaps between the local target images and the unaligned images are $6\times$ in the multiscale gigapixel dataset~\cite{yuan2017multiscale} (top of Fig.~\ref{fig:compare_giga}) and $10\times$ in the cross-resolution stereo dataset~\cite{zhou2020cross} (bottom of Fig.~\ref{fig:compare_giga}).
The results show that the conventional SIFT+RANSAC~\cite{lowe2004distinctive} fails to estimate reasonable homography matrix in the first case, and the deep learning-based methods DHN~\cite{DeTone.2016deephomography}, UDHN~\cite{Zhang.2019contentdeephomoraphy} and MHN~\cite{Le.2020dynamic} appear different degrees of missing alignments (please zoom-in for details). The proposed LocalTrans demonstrates the best visual results on the optical zoom-in cross-resolution datasets~\cite{yuan2017multiscale,zhou2020cross}, which has more complicated degradation model than synthesized cross-resolution data. More visual results on the optical zoom-in cross-resolution dataset are provided in the supplementary.

\vspace{-2mm}
\section{Conclusions}
In this paper, we proposed a novel multiscale local transformer network, termed LocalTrans, for addressing the cross-resolution problem in homography estimation. We consider the cross-resolution images as some kind of multimodal input, and employ the transformer structure to explicitly capture correspondences between two modalities (cross-resolution images) in the feature space. To accelerate the transformer, we design a Local Attention Kernel (LAK) that generates a local attention map specifically for each position in the feature. By embedding the LAK within a multiscale structure, the proposed LocalTrans is able to capture correspondences in a long-short range manner. The proposed LocalTrans network outperforms state-of-the-art methods on the MS-COCO dataset and highlights a superior performance on the challenging real-captured cross-resolution dataset under resolution gap up to $10\times$. 

We believe our LocalTrans gives a new opportunity to learn robust and accurate interactions between cross-resolution inputs, and would be further applied to various applications, such as reference-based super-resolution, cross-resolution stereo matching and hybrid light field imaging.

{\small
\bibliographystyle{ieee_fullname}
\bibliography{egbib}
}

\end{document}